\documentclass[conference]{IEEEtran}
\usepackage{graphicx,algorithm,algorithmic}
\usepackage{url}

\ifCLASSINFOpdf
\else
\fi
\renewcommand{\deg}{\theta}
\newcommand{\cp}{curvature point}
\newcommand{\cps}{curvature points}
\newcommand{\noofprim}{50}
\newcommand{\primitives}{{\em primitives}}
\newcommand{\primitive}{{\em primitive}}

\begin{document}
%
\title{Online Handwritten Devanagari Stroke Recognition Using
Extended Directional Features}

\author{\IEEEauthorblockN{Lajish VL}
\IEEEauthorblockA{
Department of Computer Science \\
University of Calicut,\\
Kerala - 673635, INDIA\\
E-mail:  lajish@uoc.ac.in 
}
\and
\IEEEauthorblockN{Sunil Kumar Kopparapu}
\IEEEauthorblockA{
TCS Innovations Lab - Mumbai\\
Tata Consultancy Services Limited\\
Thane (West), Maharashtra, India\\
Email: SunilKumar.Kopparapu@TCS.Com}
}


%


\maketitle

\begin{abstract}
This paper describes a new feature set, 
called the extended directional features
(EDF) for use in the recognition of
online handwritten strokes. 
We use EDF specifically to recognize strokes that form a basis for producing 
Devanagari  script, which is the most widely used Indian language
script. It should be noted that stroke recognition in handwritten script is 
equivalent to phoneme recognition in speech signals and is generally very poor 
and of the
order of $20$\% for singing voice. 
Experiments are
conducted for the automatic recognition of isolated handwritten
strokes. 
Initially we describe the
proposed feature set, namely EDF
and
then show how this feature can be effectively utilized for writer
independent script recognition through stroke recognition. 
Experimental results show that the extended
directional feature set performs well with about $65+$\%
stroke level recognition accuracy for writer independent data set.
\end{abstract}


%
\IEEEpeerreviewmaketitle

\section{Introduction}
\label{sec:intro}

\begin{figure}
\centering
\includegraphics[width=0.50\textwidth]{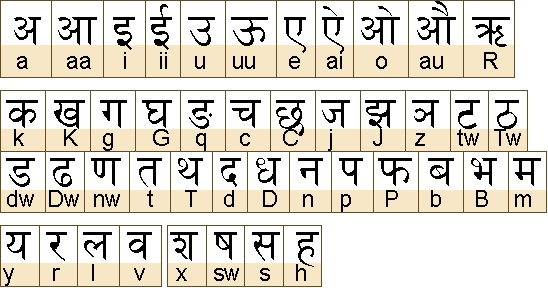}
\caption{Devanagari alphabet set \cite{web:alphabet}.}
\label{fig:alphabets}
\end{figure}

Interest in handwritten script recognition
\cite{Impedovo:2014:MTY:2562363.2562984,Cheriet:2007:CRS:1212762,Plamondon:2000:OOH:331097.331275} and
specifically in online handwritten script recognition
\cite{Kunwar:2011:OHR:2066306.2067301,Rajkumar:2012:TSO:2471876.2471996,Arora:2010:HMR:1933305.1934214} has been active for a 
long time. In the case of Indian languages, research works are active 
especially for Devanagari \cite{bb98394}, Bangla 
\cite{4761835,bb98378}, Telugu \cite{DBLP:conf/icdar/BabuPSRB07} and Tamil 
\cite{bb98368,bb98374}. Devanagari script, the most widely used 
Indian script, consists of vowels and consonants as shown in Fig.
\ref{fig:alphabets}. 
It is used 
as the writing system for over $28$ languages including Sanskrit, Hindi, 
Kashmiri, Marathi and Nepali and used by more than $500$ million people 
world wide. Devanagari is written from left to right in horizontal lines 
and the writing system is alphasyllabary. Barring a few alphabets, 
almost all the alphabets in English can be written in a single
stroke\footnote{A stroke is defined as the resulting
trace between a pen-down and its adjacent pen-up} or 
two. 
In contrast, in most Indian
languages, alphabets are made up of two or more strokes. This writing
requirement makes it
necessary to analyze a
 sequence of adjacent strokes to identify an alphabet.
Majority of the alphabets in Devanagari script  are formed by using 
multiple strokes. Language syllables are composed of vowels, 
consonants and their combinations. In a consonant-vowel combination, the 
vowels are orthographically indicated by signs called {\em matras}. These 
modifier symbols are normally attached to the top, bottom, left or right 
of the base character. Hence the consonants, vowels, matras and 
consonant/vowel modifiers constitute the entire alphabet set. These 
composite characters are then joined together by a horizontal line, 
called {\em shirorekha}. 

\begin{figure}
\centerline{ 
\includegraphics[width=0.4\textwidth]{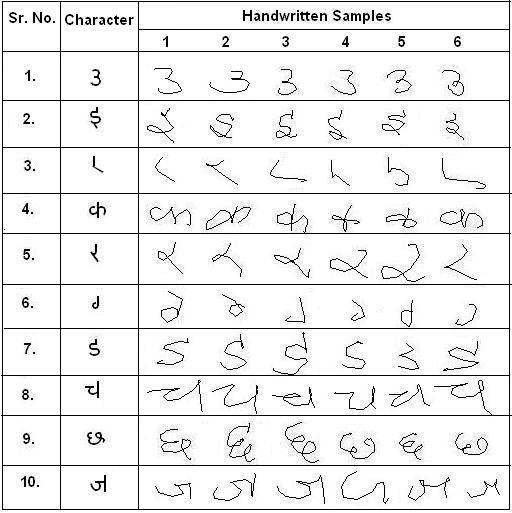}}
\caption{Sample set of \primitives\ collected from a single writer.}
\label{fig:strokes}
\end{figure}

A careful analysis based on clustering of handwritten Devanagari script 
showed that there
was    
a {\em basis like} set of 
$\noofprim$ strokes that 
was sufficient to represent 
all the alphabets in the Devanagari script.
We name these strokes \primitives. 
\begin{figure}
\centerline{ 
\includegraphics[width=0.45\textwidth]{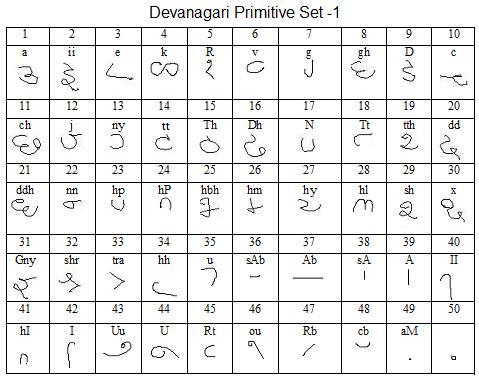}}
\caption{A set of \primitives\ hand written 
strokes that can be used to write the complete
alphabet set in Devanagari.}
\label{fig:primitives}
\end{figure}
The identified set of primitives (shown in Figure \ref{fig:primitives}) 
can be used to write the complete Devanagari alphabet set 
(Figure \ref{fig:alphabets}). 
In this paper we
use these \primitives\ as the units for recognition taking parallel from
the recognition of {\em phone set} used in speech recognition literature.
In an 
unconstrained handwriting these \primitive\ strokes exhibit large 
variability in shape, direction and order of writing. It is also 
observed that the \primitives\ are combined and broken based on the 
writer's style of writing which is acquired at the time of learning the 
script.
A sample set of \primitives\ collected from the same  
writer at different times over a period of time  
is shown in Figure \ref{fig:strokes}. The variations within the 
\primitives\ even for the {\em same} writer is evident and it is observed 
that the
variation among different writers is even larger; making the task of
recognizing these \primitives\ challenging.



\begin{figure}
\centerline{ 
\includegraphics[width=0.25\textwidth]{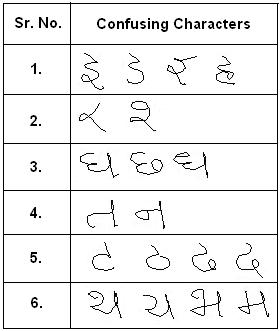}}
\caption{A list of some confusing alphabets in Devanagari.}
\label{fig:conf_strokes}
\end{figure}

While a large amount of literature is available for online handwriting 
recognition of English, Chinese and Japanese languages, until recently, 
relatively very 
less work has been reported for the recognition of Indian languages
\cite{Arora:2010:HMR:1933305.1934214,Rajkumar:2012:TSO:2471876.2471996,Kunwar:2011:OHR:2066306.2067301}. 
Even among the Indian scripts, notable work has been reported only for 
Devanagari \cite{DBLP:conf/icann/SwethalakshmiSC07}, Bangla \cite{4761835},
Tamil and Telugu scripts
\cite{conf/icdar/PrasanthBSRM07,DBLP:conf/icdar/BabuPSRB07}. 
It is also observed that 
the work done on one Indian language script cannot be directly applied 
for the recognition of a second language script 
because of the vast variation in the scripts. The 
main challenge in online handwritten character recognition in Indian 
language is the large size of the character set, variation in writing 
style (when the same stroke is written by different writers or the same 
writer at different times) and the visual similarity between different 
alphabets in the script.
A list of visually similar alphabets in Devanagari script are shown in Figure 
\ref{fig:conf_strokes}.

In this paper, we propose the use of extended directional feature (EDF) set for
the recognition of \primitives\ (which are also called strokes). 
The variations that
exist in the \primitives\ (see Figure \ref{fig:strokes}) test the 
credibility of the proposed features. 
The motivation to look at recognition of strokes rather that looking at 
alphabet recognition is influenced by the
speech recognition literature. The strokes are analogous to phonemes in
speech. It is well know in speech literature that though the phoneme 
recognition accuracies are poor (it is about $20\%$ in singing voice
\cite{DBLP:conf/icassp/MesarosV10}), 
the final output of the speech recognition is
significantly high. The poor phoneme recognition in speech recognition is
enhanced by lexicons and
statistical language models. We believe that even a poor stroke recognition
accuracies can lead to very high alphabet recognition accuracies when knowledge
about the written language is exploited.
The rest of the paper is organized as follows. 
We introduce the extended directional
feature set in Section \ref{sec:edf}, in addition a detailed explanation of 
data collection, 
pre-processing. 
Experimental results are outlined in Section \ref{sec:experimental_results},
and conclusions in Section \ref{sec:conslusions}.


\section{Extended Directional Feature Extraction}
\label{sec:edf}

Several temporal features \cite{Bahlmann2006115,Uchida:2010:ALF:1904935.1905964} have been used for script recognition in 
general and for online Devanagari script \cite{5430176} recognition in particular. We 
propose a simple yet effective feature set based on extended 
directional chain 
code. The detailed procedure for obtaining these directional 
features is given below.

\begin{figure}
\centerline{ 
\includegraphics[width=0.35\textwidth]{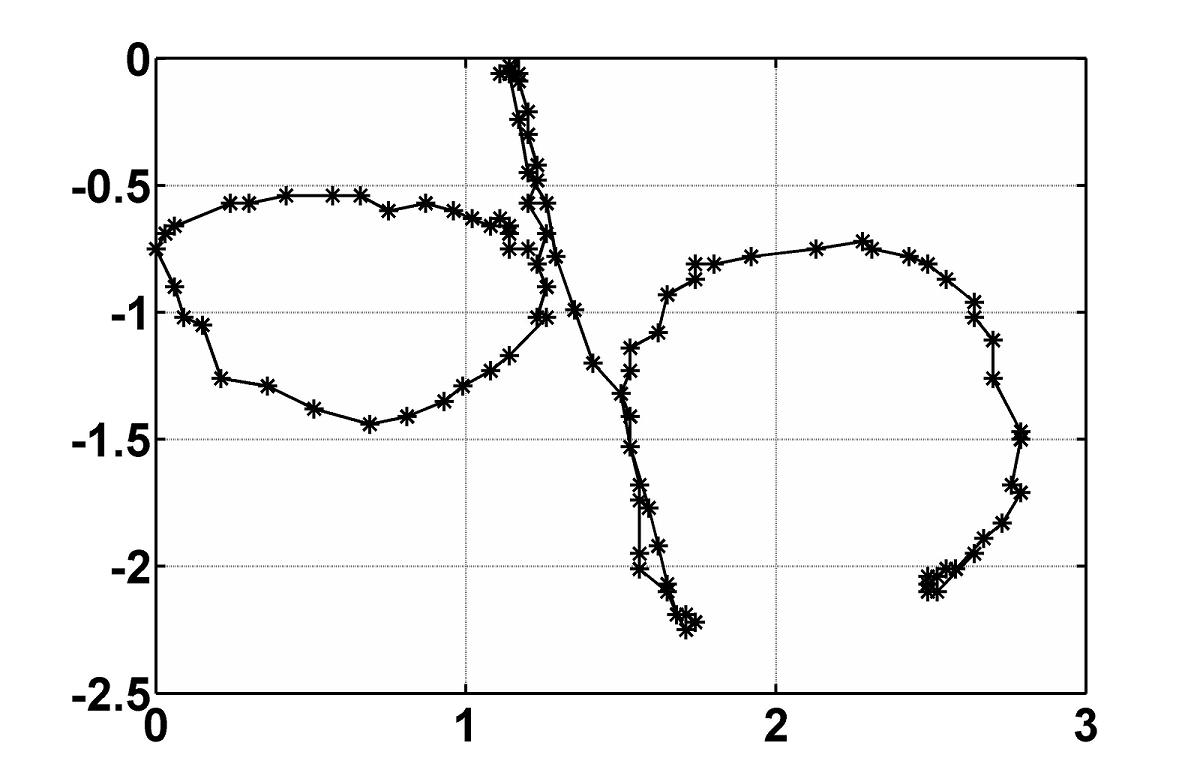} 
}
\caption{
A sample online character. The "*" represent the $(x,y)$ points,
the points have been joined to give a feel for the stroke.} 
\label{fig:sample}
\end{figure}

Let the online stroke be represented by a variable number of 2D 
points which are in a time sequence. 
For example an online stroke would 
be represented as
 \[
\left \{(x_{t_1} , y_{t_1} ), (x_{t_2} , y_{t_2} ), \cdots , (x_{t_n} ,
y_{t_n}) \right \}
 \] where, $t$ denotes the time and assume that $t_1 < t_2 < \cdots · · 
< t_n$. Equivalently we can represent the online stroke (see Figure 
\ref{fig:sample}) as \[ \left \{(x_1 , y_1 ), (x_2 , y_2 ), \cdots , 
(x_n , y_n ) \right \} \] by dropping the variable $t$. The number of 
points denoted by $n$ vary depending on the size of the stroke and 
also the speed with which it was written. Most handwritten script 
digitizing devices 
(popularly called electronic pen or e-pen) 
sample the handwritten stroke uniformly in time. For this 
reason, the number of points per unit length of a handwritten stroke 
is large when the writing speed is 
slow which is especially true at curvatures (see Figure \ref{fig:sample}) and
vice-versa.

\begin{figure}
\centerline{ 
\includegraphics[width=0.35\textwidth]{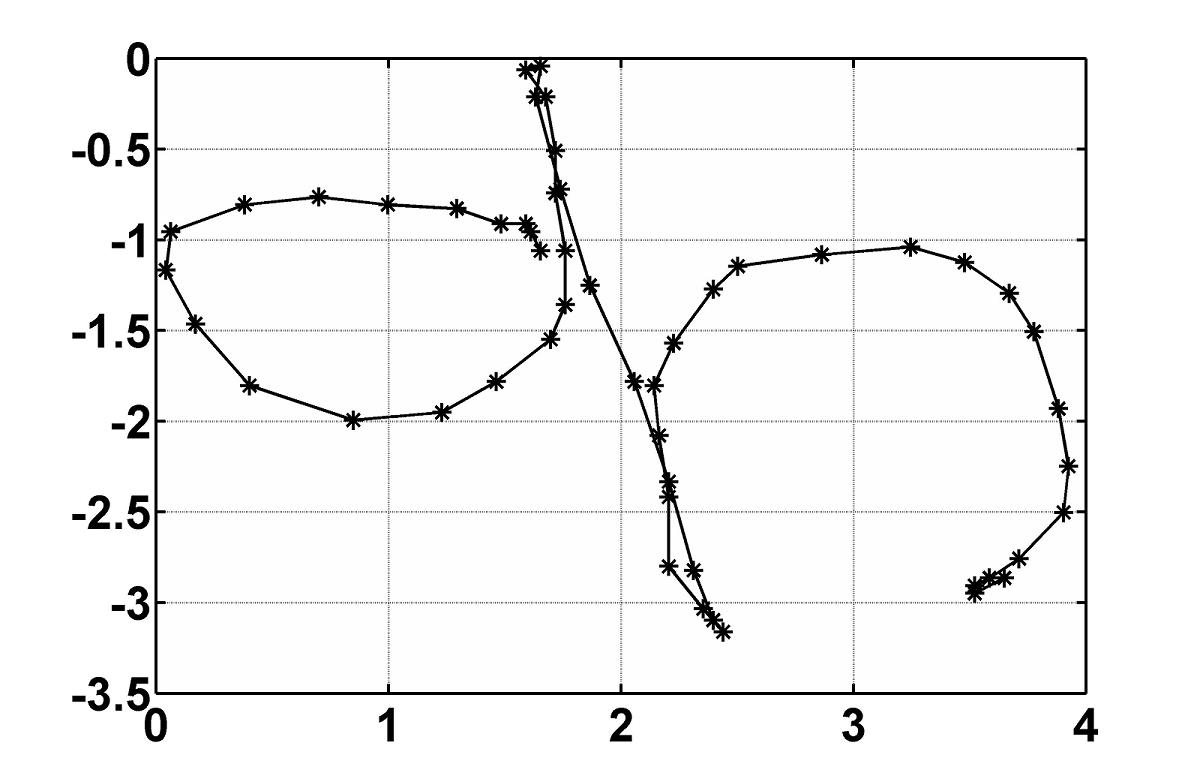}}
\caption{
After smoothing Figure \ref{fig:sample} using Discrete Wavelet transform.
}
\label{fig:dwt}
\end{figure}

We first identify the curvature 
points (also called critical points) from the smoothed (we use discrete
wavelet transform, we could have used any noise removal technique
\cite{5697488}, see Figure \ref{fig:dwt}) handwriting data. The sequence $(x_i, y_i)_{i=0}^{n}$
represents the handwriting data of a stroke. We treat the sequence $x_i$ and
$y_i$ independently and compute the \cps\ for each of these sequence. 
For the $x$ sequence, we calculate the first difference 
\[x'_i = sgn(x_i - x_{i+1})\]
where
\begin{eqnarray*}
sgn(k) = +1 &\mbox{if} & x_i - x_{i+1} >0\\ \nonumber
sgn(k) = -1 & \mbox{if} & x_i - x_{i+1} <0 \\ \nonumber
sgn(k) = 0 & \mbox{if} &x_i - x_{i+1} =0 
\end{eqnarray*}
\begin{figure}
\centerline{ \hfill
\includegraphics[width=0.23\textwidth]{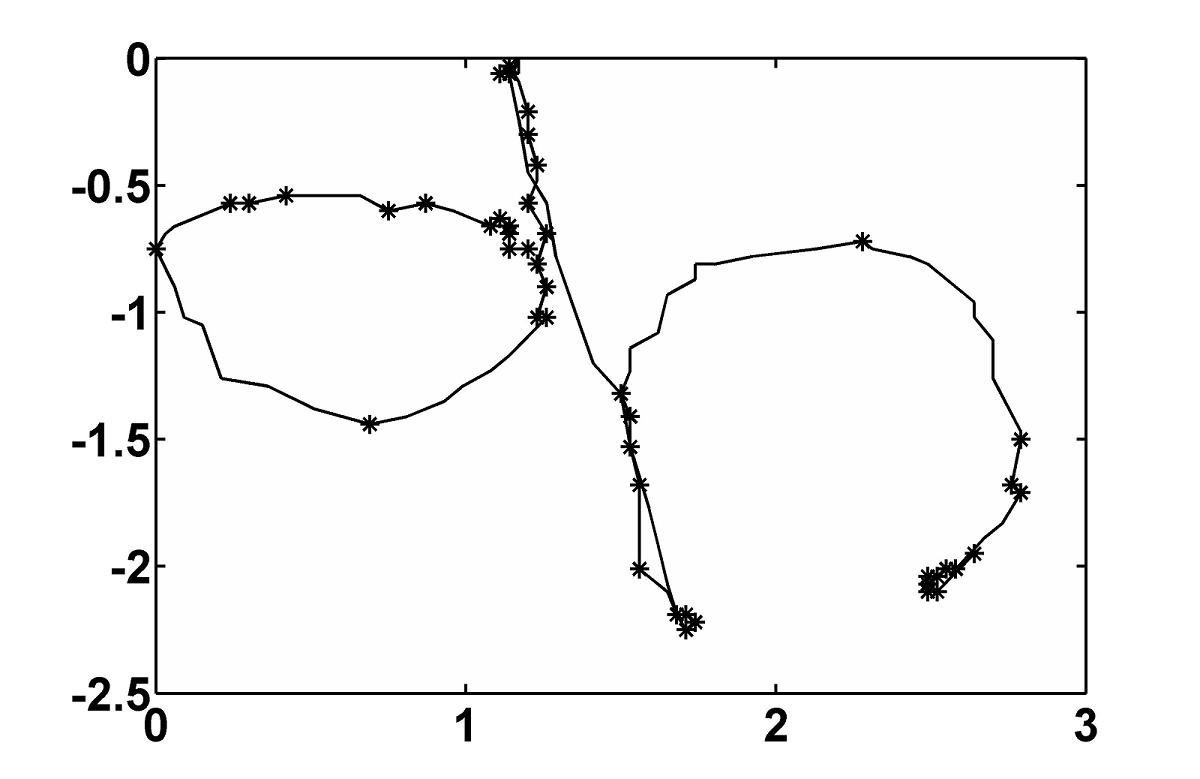} \hfill
\includegraphics[width=0.23\textwidth]{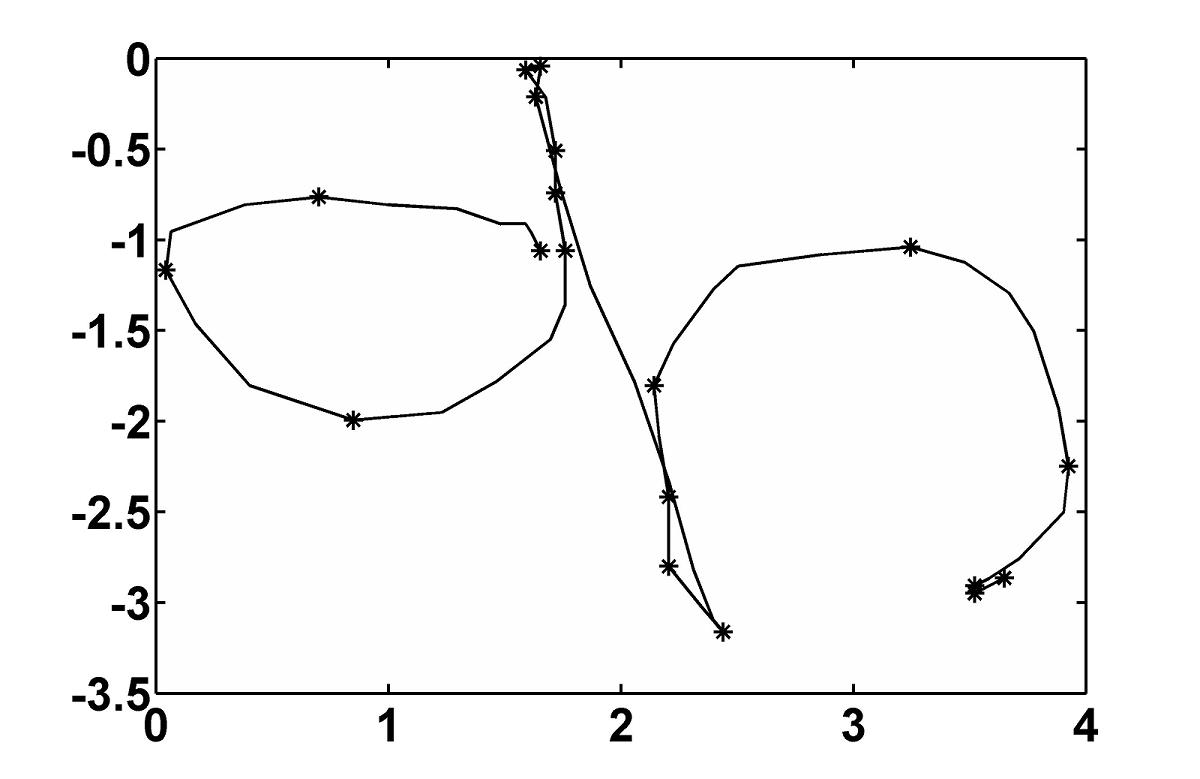}\hfill}
{\hfill (a) \hfill (b) \hfill}
\caption{Curvature Point extraction. (a) Raw online character and (b) after smoothing using Discrete
Wavelet transform.
}
\label{fig:critical_points}
\end{figure}
%
We use $x'$ to compute the \cp. The point $i$ is a \cp\ {\em iff}  \[ x'_i - x'_{i+1} \ne 0. \]
Similarly we calculate the \cps\ for the $y$ sequence. The final list
of \cps\ is the union of all the points marked as \cps\ by
both the $x$ and the $y$ sequence (see Figure \ref{fig:critical_points}).
Clearly the
number and position of the  \cps\  are more consistent and occur 
at the points
where there is  a change in curvature for smoothened stroke (Figure
\ref{fig:critical_points}(b)) when compared to a raw stroke (see Figure
\ref{fig:critical_points}(a)).
It must be noted that  the position and number of curvature points computed for
different
samples of the same stroke may vary.

Let $k$ be the number of curvature points
(denoted by $c_1, c_2, \cdots c_k$) extracted from a stroke of length $n$; usually $k << n$.
The $k$ \cps\ form the basis for extraction of the extended
directional features.
We first compute the angle between the two \cps,
say $c_l$ and $c_m$, 
as \[
\theta_{lm} = \tan^{-1}\left (\frac{y_l - y_m}{x_l - x_m} \right ) \]
Now the
extended 
directional feature set is computed by computing the direction between the
\cps\ as shown in Figure \ref{fig:edf}. Where $d_{lm}$ 
corresponding to the angle $\theta_{lm}$ (computed using the Algorithm
\ref{algo:cp2dir}) and is the direction between
the \cp\  $c_l$ and $c_m$.


\begin{algorithm}
\begin{algorithmic}
\STATE{int deg2dir(double $\deg$)}
\STATE{int dir = -1;}
\IF{($\deg > -\pi/8$ \& $\deg < \pi/8$)}
	\STATE{dir = 1;}
\ENDIF
\IF{($deg >= \pi/8$ \&  $\deg < 3 \pi/8$) }
	\STATE{dir = 2;}
\ENDIF
\IF{($\deg >= 3 pi/8$ \&  $\deg < 5 pi/8$) }
	\STATE{dir = 3;}
\ENDIF
\IF{($\deg >= 5 \pi/8$ \& $\deg < 7 \pi/8$) }
	\STATE{dir = 4;}
\ENDIF
\IF{(($\deg >= \frac{7 \pi}{8}$ \& $\deg < \frac{9 \pi}{8}$) 
$\|$ ($\deg >= -\frac{9 \pi}{8}$ \& $\deg
< - \frac{7 \pi}{8}$))}
	\STATE{dir = 5;}
\ENDIF
\IF {($\deg >= -7 \pi/8$ \& $\deg < -5 \pi/8$) }
	\STATE{dir = 6;}
\ENDIF
\IF {($\deg >= -5 \pi/8$ \& $\deg < -3 \pi/8$) }
	\STATE{dir = 7;}
\ENDIF
\IF{($\deg > -3 \pi/8$ \& $\deg < -\pi/8$) }
	\STATE{dir = 8;}
\ENDIF
\STATE{return(dir);}
\end{algorithmic}
\caption{Angle between two \cp\ conversion into direction}
\label{algo:cp2dir}
\end{algorithm}
\begin{figure}
$
\begin{array}{cccccccccc}
\mbox{CP} & c_1 & c_2 & c_3 & c_4 & \cdots& \cdots & c_m &\cdots &  c_k \\ \hline
c_1&0&d_{12}&d_{13}&d_{14}&\cdots&\cdots&d_{1m}&\cdots&d_{1k} \\
c_2&-&0&d_{23}&d_{24}&\cdots&\cdots&d_{2m}&\cdots& d_{2k}\\
c_3&-&-&0&d_{34}&\cdots&\cdots&d_{3m}&\cdots& d_{3k}\\
c_4&-&-&-&0&&&&& d_{4k}\\
\vdots &\vdots&\vdots&\vdots&\vdots&\vdots&\vdots&\vdots&\vdots&\vdots \\
c_l&-&-&-&-&&&d_{lm}&& d_{lk}\\
\vdots &&&&&&&&& \\
c_k&-&-&-&-&-&-&-& -&0\\
\end{array}
$
\caption{Extended Directional Features}
\label{fig:edf}
\end{figure}
Given, $k$ \cps, we get (see Figure \ref{fig:edf}) an extended directional feature (EDF) vector of size 
\[
\frac{k(k-1)}{2}.
\]
In all our experiments we have used this extended directional feature set to
represent a stroke.

\section{Experimental Analysis}
\label{sec:experimental_results}

\begin{figure}
\centerline{ 
\includegraphics[width=0.47\textwidth]{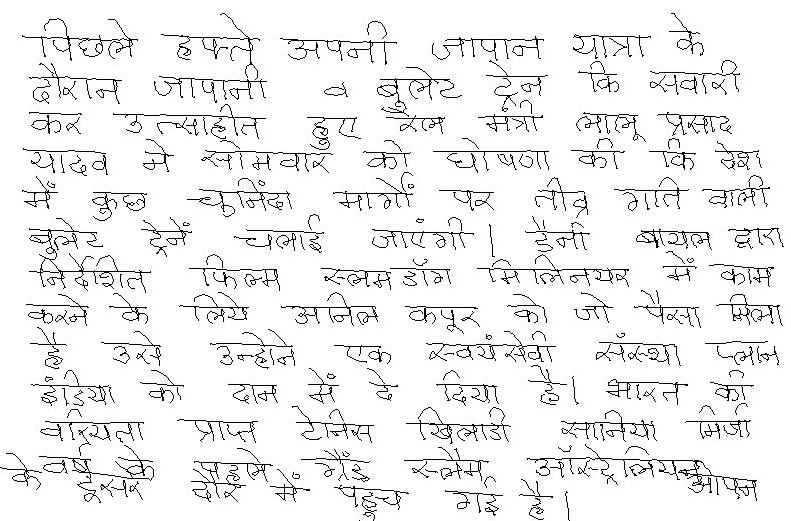}
}
\caption{Paragraph of online data collected from a user.}
\label{fig:sample_data}
\end{figure}

For experimental analysis, we collected data from $10$ persons, each of whom
wrote a paragraph of Hindi text using 
Mobile e-Notes Taker
(see for example, Figure \ref{fig:sample_data}). 
The mobile e-note taker is a portable pen based handwriting 
capture device which allows user to write on a 
normal paper using the electronic pen to capture the online
handwritten text. 
The SDK provided with 
the device enables extraction of the $x,y$ trace of the  online 
handwriting data. In addition to the $x,y$ trace the pen captures 
the pen-up and
pen-down information which helps identify a stroke. 
Each stroke is characterized by a $x,y$ sequence between
a pen-down and a pen-up point. This raw stroke level data is smoothed
using Discrete Wavelet 
Transform (DWT) decomposition, 
as mentioned earlier we do not dwell on this in this paper
since this is well covered in pattern recognition literature, 
to remove noise in terms of small undulation due
to the sensitiveness of the sensors on the electronic pen. 
For each stroke we extracted the extended directional feature set as described
in Section \ref{sec:edf}.

We used $5$ user paragraph data for training and the other (not part of
training) $5$ for the purpose 
of testing
the performance of the ED feature set. 
We constructed a total of $252$ ($^{10}C_{5}$) sets of training and test data.
We initially hand tagged each stroke
in the collected data using the $\noofprim$ \primitives\ that we selected (see Figure
\ref{fig:primitives}). The strokes
that did not fall into this \primitive\ set were marked as being out of
vocabulary. All the strokes corresponding to the given \primitive\ in the
training data was collected and clustered together. We retained those
\primitives\ that occurred atleast $10$ times in the train and the test data 
and the rest of the \primitive\ were not used for training and testing.
In all we were able to get $20$ \primitives\ which occurred atleast $10$ times in
both the training and the test data set. While the dataset is not very large,
the $252$ different runs demonstrates the 
effectiveness of EDF in recognition of the \primitives.

As a next training step, we calculated the 
dynamic time warping (DTW) distance between all strokes corresponding to the same
\primitive\ (numbering $20$).  
Note that different strokes corresponding to the same \primitive\ had
different ED length and hence to compute the distance between the two strokes
we need to use DTW algorithm\footnote{We do not discuss this
algorithm, in detail since it is well used in online script recognition
literature \cite{Bahlmann:2004:WIO:968725.968862}.}. All strokes corresponding to the
same \primitive\ which were within a distance of $\tau$ were clustered together 
and only one representative stroke 
from the cluster was retained as the cluster representative. We
chose $\tau$ such that for each \primitive\ there were a maximum of $3$ sample
strokes. So for a set of $20$ \primitives\ we had a reference set of $60$
samples.

For testing purpose, we took a test stroke ($s_t$) from the test data, 
we first
extracted EDF and compared it with
the EDF of the $60$ reference strokes using DTW
algorithm. We choose $2$ different methods to assign the test stroke 
into one of the $20$ \primitives\ (classification). 
\begin{itemize}
\item{Method I}: 
The stroke $s_t$ is classified as a \primitive\ $p*$ such that the DTW distance
of $s_t$ with the \primitive\ $p*$ is minimum
\[
\min_{p*} \left \{{d(s_t,p_i) }_{i=1}^{60} \right \}
\]
Note $d(a,b)$ is the DTW distance between $a$ and $b$.
\item Method II: The stroke $s_t$ is compared with all the $60$ reference
strokes and
the distance $d(s_t, p_i)$ for $i=1, \cdots 60$ computed. We then take the average distance of the stroke from all
the $3$ references of a \primitive. We arrange these average distances ($20$ in
number) in the increasing order of magnitude. The \primitive\ with the least
average distance from the test stroke $s_t$ is declared as being recognition of
stroke $s_t$.
\end{itemize}

\begin{table}
\caption{Average (over $252$ runs) number of strokes under each \primitive\ in the test data set.}
\label{tab:data}
\begin{center}
\begin{tabular}{||l|c||} \hline
Primitive &  Test \\ \hline
R & 71 \\
l & 18 \\
k & 57 \\
nn & 67 \\
v & 36 \\
p & 36 \\
dd & 50 \\
m & 50 \\
tt & 23 \\
y & 36 \\
aa & 28 \\
h &16 \\
T & 25 \\
g &25 \\
D & 21 \\
e & 16 \\
j & 18 \\
ii & 10 \\
ch & 11 \\
c & 11 \\ \hline
Total & 625 \\ \hline
\end{tabular}
\end{center}
\end{table}

Table \ref{tab:data} shows the average number of strokes in the test data set
corresponding to the $20$ selected \primitives.  
All the experimental results are based
on this data set (from $10$ people but run $252$ times). 
The overall average stroke level recognition
accuracies for both Method I and Method II did not vary significantly and stood
at $65.6$ \% and $65.9$ \% respectively. Meanings on an average 
$410$ for Method I and $412$
for Method II of the $625$ strokes were correctly recognized. Details of the
average recognition are captured in Table \ref{tab:results}. 
It should be noted that
the accuracies are writer independent and for stroke level recognition.

\begin{table}
\caption{Average (over $252$ runs) Recognition Accuracies for test data set.}
\label{tab:results}
\begin{center}
\begin{tabular}{||l|c|c|c||} \hline
Primitive &  \# Test  & \# recognized & \# recognized\\ 
 &    & Method I & Method II\\ \hline
R & 71 & 46& 40\\
l & 18 & 7& 9\\
k & 57 & 51& 53\\
nn & 67 & 36& 34\\
v & 36 & 23& 26\\
p & 36 & 32& 32\\
dd & 50 & 34& 36\\
m & 50 & 36& 40\\
tt & 23 & 11& 13\\
y & 36 & 22& 20\\
aa & 28 & 25& 20\\
h &16 & 9& 9\\
T & 25 & 15& 16\\
g &25 & 10& 12\\
D & 21 & 13& 15\\
e & 16 & 7& 5\\
j & 18 & 13& 13\\
ii & 10 & 4& 5\\
ch & 11 & 9& 8\\
c & 11 & 7& 6\\ \hline
Total & 625 & 410& 412\\ \hline
\end{tabular}
\end{center}
\end{table}

\section{Conclusions}
\label{sec:conslusions}

In this paper we have introduced a new online feature set, 
called the extended directional feature. Based on the extensive experimentation
($252$ runs) on a set of strokes captured from a set of $10$ people, we observe
that this feature set 
is capable of discriminating similar looking strokes quite well. We have
presented recognition accuracies for writer independent stroke level data set. 
It is
well known, both in speech and script recognition literature that stroke
(phoneme in case of speech) recognition is always poor. However like in speech
where the phone recognition is improved by using lexicon and statistical
language model, we plan
to cluster strokes using spatio-temporal information to form alphabets and then
use the cluster of strokes to classify them into an alphabet. This we believe
will lead to good accuracies of writer independent script recognition. Further
the derived \primitives\ set are not language dependent and can be used for 
recognition of other languages albeit with a different \primitive\ set. 


\bibliographystyle{IEEEbib}
\bibliography{script_stat_analysis} 

%
%
\end{document}